\documentclass[letter,12pt]{article}
\usepackage[margin=25mm]{geometry}
\usepackage{adjustbox}
\usepackage{amsmath}
\usepackage{amsfonts}
\usepackage{amssymb}
\usepackage{graphicx}
\usepackage{makecell}
\usepackage{verbatim}
\immediate\write18{texcount -tex -sum  \jobname.tex > \jobname.wordcount.tex}

\title{A Review on Generative Adversarial Networks for Data Augmentation in Person Re-Identification Systems}
\author{V\'ictor Uc-Cetina$^{1}$, Laura \'Alvarez-Gonz\'alez$^{2}$, Anabel Martin-Gonzalez$^{1}$, \\
\small $^{1}$ Universidad Aut\'onoma de Yucat\'an - \small \{uccetina, amarting\}@correo.uady.mx \\
\small $^{2}$ Universidad Aut\'onoma de Yucat\'an - \small laura.alvargonza@gmail.com \\
}
\date{February 2023}

\begin{document}

\maketitle

\begin{abstract}
Interest in automatic people re-identification systems has significantly grown in recent years, mainly for developing surveillance and smart shops software.  Due to the variability in person posture, different lighting conditions, and occluded scenarios, together with the poor quality of the images obtained by different cameras, it is currently an unsolved problem. In machine learning-based computer vision applications with reduced data sets, one possibility to improve the performance of re-identification system is through the augmentation of the set of images or videos available for training the neural models. Currently, one of the most robust ways to generate synthetic information for data augmentation, whether it is video, images or text, are the generative adversarial networks. This article reviews the most relevant recent approaches to improve the performance of person re-identification models through data augmentation, using generative adversarial networks. We focus on three categories of data augmentation approaches: style transfer, pose transfer, and random generation.
\end{abstract}

\maketitle

\section{Introduction}
Re-identification  of a person consists of the recognition of the same person through videos obtained from different cameras with a non-overlapping range of vision, captured at different moments in time. If we focus on the security domain, every year the installation of security cameras in cities around the world increases. The amount of images generated by these cameras every second translates into the need for manually, semi-automatically or automatically analysis of hundreds of hours of surveillance cameras by security forces.

\begin{figure*}
\centering
 \includegraphics[width=0.8\linewidth]{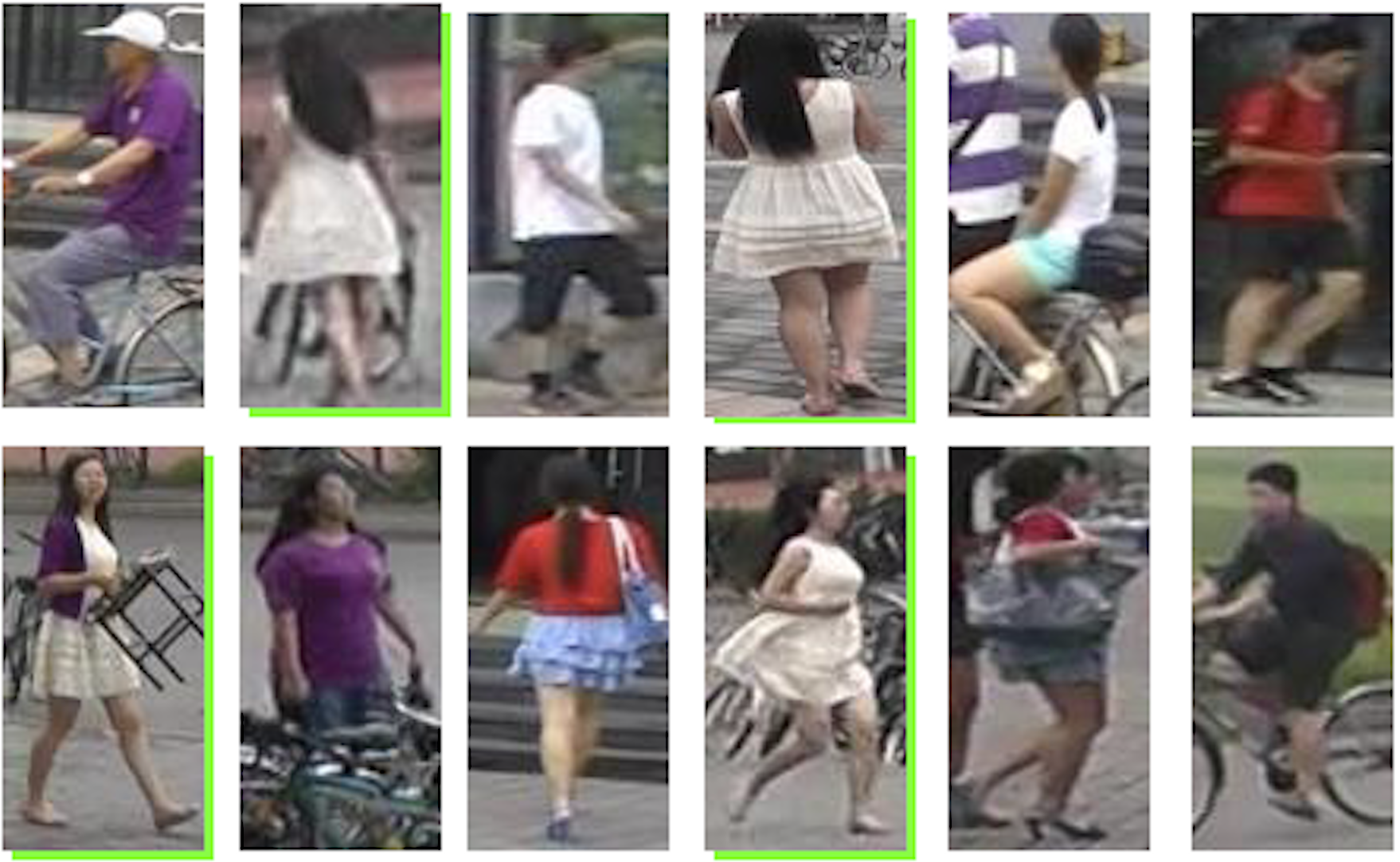}
 \caption{Example of Market-1501 \cite{MarketZheng2015} images obtained by security cameras. Images corresponding to the same person are marked with green borders. The low resolution of the images prevent us from using automatic facial recognition information.} 
 \label{fig:problemsREID}    
\end{figure*}

In recent years, interest in the problem of re-identification has increased, due to the fact that it is not possible to use biometric data for the identification of a person, since the quality of the images does not allow a clear definition of the facial data. Therefore, other image features must be used, such as the structure of the body and clothing. Currently, different neural network models have been proposed to improve the re-identification performance, but to make these models work more efficiently, they need to be trained with very diverse data. One way to artificially generate diverse data is by means of generative adversarial networks (GANs)  \cite{Goodfellow2014} which are capable of generating synthetic data that can help to improve the performance of the re-identification neural networks.

Training a basic GAN model involves two main steps:

\begin{enumerate}
\item In the first step, we train a neural network to receive as input a vector of randomly generated numbers, and produce as output the pixel values for a synthetic image. This neural network is called the generator network. At the same time, during this step, we train a second neural network, in this case, the so-called discriminator network. The discriminator receives an image as input and produce a classification as output, indicating whether the input images is a real one or a fake one.

\item In the second steps, we use the generator network to produce some fake images, but we label them as real images, and we pass them to the discriminator network. The discriminator network will produce an error signal, given that the image is fake, but it is labelled as real one. This error signal is used to further train the generator network, making it to produce more realistic fake images.
\end{enumerate}

These two steps are intertwined during several iterations until both networks improve their performance to discriminate and generate fake images. In general, It is not a trivial task to train GANs, especially considering the large amounts of data and computational power needed. However, as we describe in this article, there has been already some successful approaches.
 
On the other hand, due to the current increase in the amount of information from video surveillance systems that has to be analyzed, the implementation of an intelligent automatic analysis system is extremely important. In some countries, such as China, advanced cameras with sufficient resolution to extract unique face features are currently being used. However, since most surveillance cameras generate low-resolution images, the option of generating an automatic facial recognition system for re-identification is ruled out.

As it can be seen in Fig.~\ref{fig:problemsREID}, images are low-resolution, with variations in lighting and contrast, in addition to other factors that make the re-identification of people in images more complex, such as: the change of clothing, whether or not the person is wearing a backpack or sweater, the presence of obstacles or people in the background that limits the visibility of the person of interest in an open space.

In order to support the development of re-identification neural networks, different deep training libraries have been developed, such as Torchreid \cite{Zhou2019a}. However, to achieve effective training of these models, a large amount of correctly labeled data is needed. This constitutes one of the greatest challenges for training efficient re-identification models. Currently, the largest datasets are very limited because they do not contain a large number of images. For example, two of the most important datasets, Market1501 and DukeMTMC-reID have 1501 persons recorded on 6 different cameras and 702 persons recorded on 8 different cameras, respectively.

The purpose of this article is to review the most relevant data augmentation methods based on the use of generative adversarial networks for people re-identification, in such a way that it provides a reference guide to researchers and engineers interested in this field.

\section{MOST COMMON APPROACHES}

Generative adversarial networks, proposed in 2014 by Ian Goodfellow et al.~\cite{Goodfellow2014}, are capable of artificially generating images with great diversity. Over time, new architectures have been generated that improve the quality of the data generated, such as the CycleGan architecture, proposed in 2017 by Zhu et al. \cite{Zhu2017}. Apart from improving the quality of the generated images, it manages to transfer the style or domain of a group of images to another group, using two generative adversarial networks. 

The performance of adversarial generative networks has been improving year after year, as new models have been proposed. This improvement has motivated several researchers to start investigating the use of GANs for data augmentation tasks. An important increase in the study of GANs for data augmentation in the training of re-identification models can be noted since 2018. The most relevant approaches have been grouped into three categories, corresponding to different methods used to generate new artificial images.

\begin{enumerate}
\item Style transfer. New images are artificially generated from an input image, using different styles, known also as domains. The styles are imposed on the new images at the moment they are generated by neural networks previously trained for that purpose. In the new generated images, you can see modifications with respect to the input image, such as color, tone, and lighting.

\item Pose transfer. In this approach the inputs are one image of a real person and the target posture that we want to impose on that person. The posture can be specified whether as a heat map or by the joints that correspond to the skeleton of the desired posture. The model is capable of generating the image of the input person with the determined posture.

\item Random generation. As its name suggests, methods in this category are less constrained and they are focused on randomly generating synthetic images with the only condition that the generated images should have similar characteristics of those images in the dataset that we want to augment. 
\end{enumerate}

\section{STYLE TRANSFER}
Images obtained by cameras usually have different resolutions or are in different positions. This can cause the lighting and tonality to vary, among other aspects. One way to generate new data is by transferring from one domain to another or adapting the domain, which is based on the idea of transferring the style of one or more images to others without changing the structure nor the background of the original images. Which means that there is no change in the positions of the pixels, leaving the images exactly the same.

There are several ways of dealing with style transfer. CycleGAN \cite{Zhu2017} is a GAN introduced in 2017. It is capable of learning the style of some images and transferring it to other different images, that is, transferring the style from one domain to another. This was a milestone within the generative adversarial networks and as of 2018 a large number of works based on this architecture began to appear. For instance in 2019, Zhong et al. \cite{Zhong2019} proposed CamStyle for transferring the style of a security camera to that of another camera. The newly proposed method based on CycleGAN can transfer the style only between two domains, limiting the architecture in such a way that it is necessary to generate a model for each pair of security cameras.

Following the same philosophy of transferring one style from another, Dai et al. \cite{Dai2018a} proposed the cmGAN model focused on style transfer to convert RGB and infrared camera images. It is the first method using a RGB-Infrared Cross-Modality Re-ID Dataset which includes images from four infrared and two RGB cameras. In this case, the generative adversarial network's discriminator is part of the feature extractor of the re-identification model. The input to the re-identification model is an infrared image and the target person needs to be searched within the RGB images.

All these methods mentioned so far have the limitation of only transferring a style from one domain A to another domain B, which results in the need to duplicate the project for each different style. Furthermore, one of the biggest challenges faced by the models of re-identification is the poor performance obtained when images from one database are used during the testing of a re-identification model trained with images from another database. Therefore, some works propose the transfer of domains between different databases and/or multiple domains. 

In 2018, as an improvement to the Camstyle architecture and seeking better performance using the model in different databases, the M2M-GAN \cite{Liang2018} architecture is proposed, which classifies the images of each database into subdomains, that is, for each of the cameras. Being able to transfer the sub-domain of domain A to a sub-domain of domain B, the training being carried out in a supervised manner and requiring that all the data from the different databases have been labeled by hand.

\begin{figure*}
\centering
\includegraphics[width=0.8\linewidth]{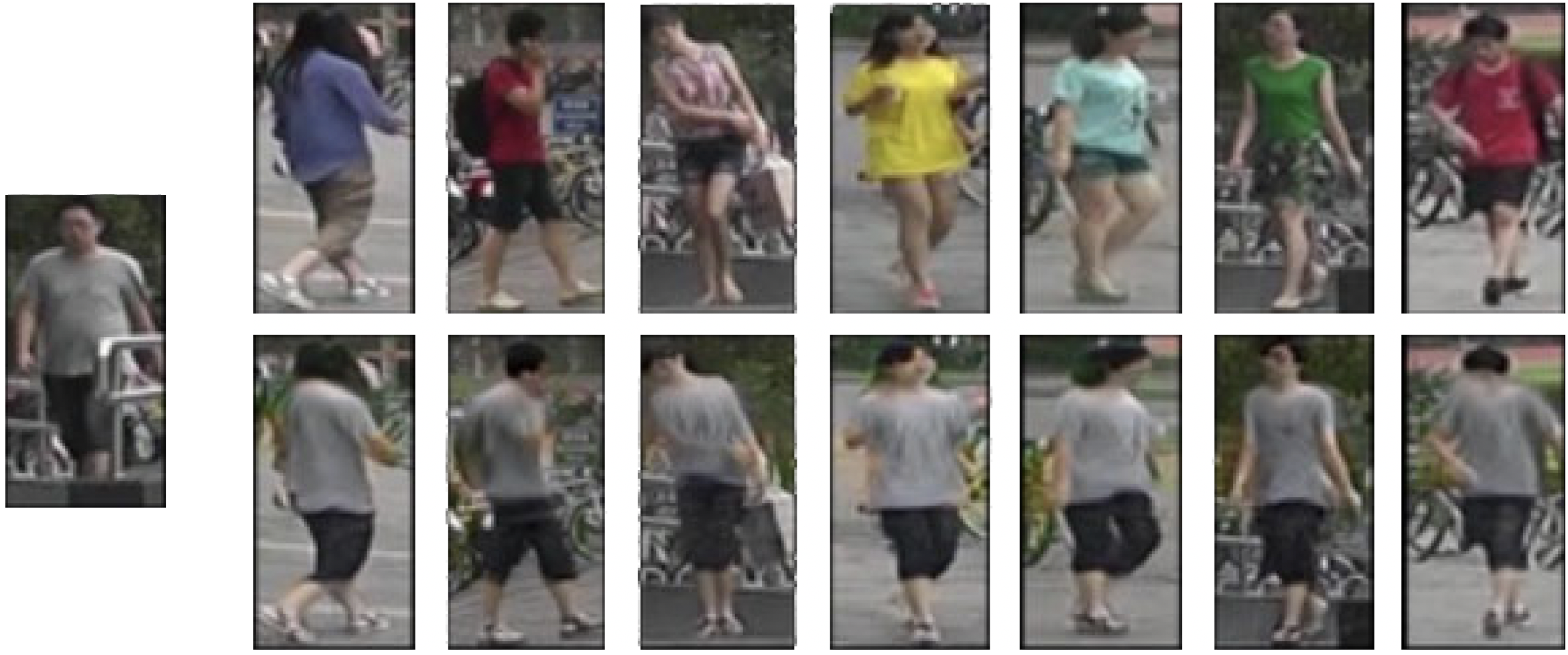} 
\caption{Style transfer. In this example of DE-GAN \cite{Zheng2019}, the algorithm transfers the appearance of the image on the left to all images on the right, combining appearance and structure.}
\label{fig:aparienciaestructura}
\end{figure*}

As of 2019, different more sophisticated architectures began to appear, such as the one proposed by Zheng et al. \cite{Zheng2019}, DG-NET, which uses two encoders that are capable of extracting the colors of the image, appearance, and transfer those colors to another image where the structure of the person has been extracted . This model also uses the adversarial generative network's own architecture as a re-identification model (see Fig. \ref{fig:aparienciaestructura}).

Some architectures can be trained to differentiate between different colors of clothing and preserves the coherence of identity of the same person with the color of clothing. Such color-aware architectures are capable of differentiating and modifying the different colors of upper clothing (sweatshirts/t-shirts) or lower clothing (pants).

Zhiqi Pang et al. \cite{Pang2021} in the year 2021 proposed a hybrid method, supervised and unsupervised. This method uses a novel architecture TC-GAN to generate labelled artificial images, transferring the person from the input image to the background of the target style image. They also proposed the use of the DFE-Net re-identification model that uses a modified version of the ResNet-50 network, pretrained with the ImageNet database that has as input the real images without tag and artificially generated ones. It uses the network as an image feature extractor for later comparison.

\section{POSE TRANSFER}
One of the biggest challenges within the re-identification problem is the great variation in the posture of a person that can be seen in different cameras. To mitigate this problem, the generation of new data of the same person is proposed, modifying the posture using different architectures. In this approach, the data augmentation process is based on the generation of new data through the extraction of the person in the original image, which can be done by obtaining joints or heat maps. The pose transfer is obtained with the help of the original image and those generated maps encapsulating the posture information.

\begin{figure*}
\centering
\includegraphics[width=0.8\linewidth]{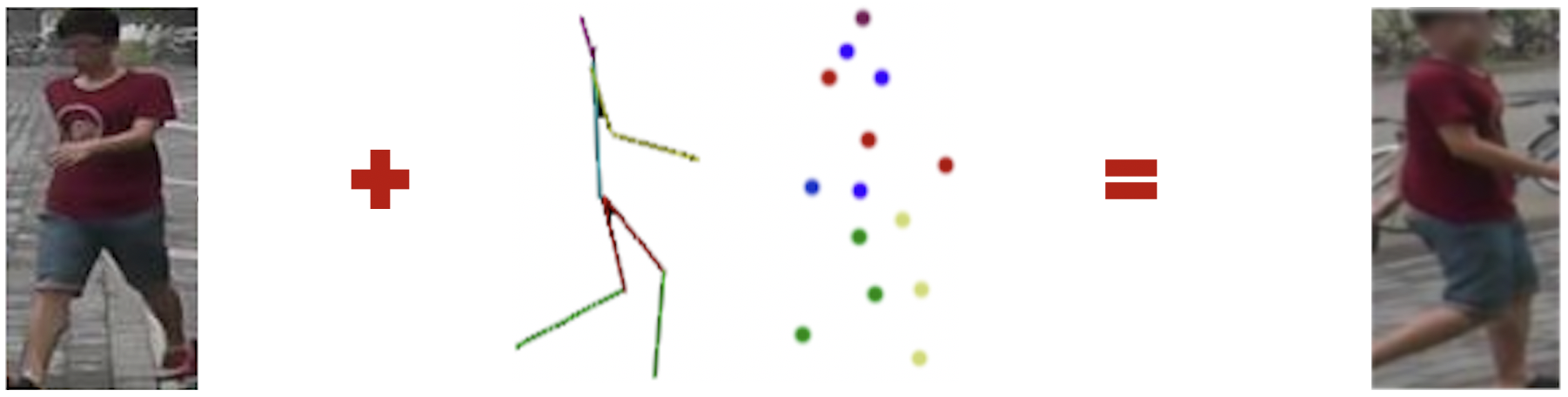} 
\caption{Pose transfer. A new image of the person is generated from an input image and the skeleton and the heat map of the desired posture.}
\label{fig:articulacionesmapas}
\end{figure*}

In 2018, Qian et al. \cite{Qian2018} proposed the PN-GAN architecture. It is capable of generating artificial images by means of the image of a person in eight different postures. The eight canonical postures are obtained using the k-means algorithm on the distribution of all the images in the database. To generate the template, the open pose \cite{Cao2017} estimator tool is used, which is capable of detecting 18 joints of the human body and their joints. Through the articulation map of both images it is able to transfer the posture of each one of the eight canonical postures to the input images, as illustrated in Fig. \ref{fig:articulacionesmapas}.

In 2019, Borgia et al. \cite{Borgia2019} following the same line as the previous architecture and extracting the joints using also the open pose method, they propose an architecture which, instead of evaluating a single image, it evaluates video footage of a person's movement. Eight canonical postures are predefined: three looking forward, three from behind, one in profile looking to the right side and another to the left side. In the first place, the video sequences of a person are obtained and the corresponding images with the canonical postures are searched by means of the Euclidean distance, if any of those eight postures does not exist, the artificial image will be generated. The same happens with all the people that are in the video, their sequences are obtained and if there is no canonical position it generates it, by means of the cosine distance they compare the eight corresponding images with the eight canonical postures of the input person with all the other sequence of images of people. A classification is made and the one with the smallest cosine distance is assigned as if it were the same person in the re-identification model.

In 2020,  Zhang et al. \cite{Zhang2020} proposed PAC-GAN. It consists of two models, the first is CPG-Net, through the use of a conditional GAN, it is capable of generating artificial images of a person from a camera A, and converting it to the point of view of another camera B. The data is increased by generating new images with postures from different points of view of different cameras. It trains with the joints generated with Open Pose and the image itself. 

More recently, in 2021, Ziyang et al. \cite{Ni2021}  proposed a new architecture which is capable of correcting images in such a way that the images of people are centered and straight. For this, the database was trained in such a way that the images with the correct positions were indicated.

\section{RANDOM GENERATION}
\begin{figure*}
\centering
\includegraphics[width=0.8\linewidth]{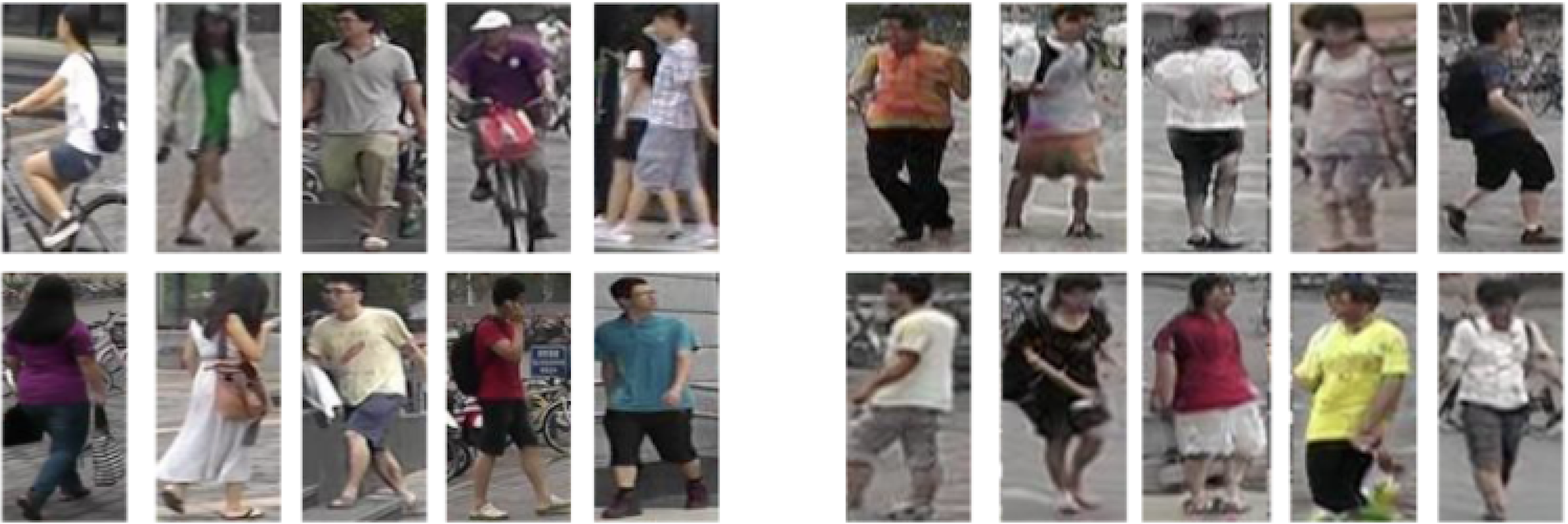} 
\caption{Random generation. On the left, real images taken from the Market-1501 \cite{MarketZheng2015} database. On the right, artificial images randomly generated with StyleGAN3 \cite{Karras2021}.}
\label{fig:imagenresrealisficticias}
\end{figure*}

The third major approach consists in generating random images of persons, with different postures, lighting, colors and backgrounds (see Fig. \ref{fig:imagenresrealisficticias}). After the images has been generated, different methods are used to automatically label them. Finally the labelled images are used for training the re-identification model.

One of the main algorithms for the automatic labeling of the random images is called Label Smooth Regularization (LSR) and it was proposed in 2015 by Szegedy et al. \cite{Szegedy2016}. This algorithm was originally used in an image classification problem.

Later on, in 2017, Zheng et al. \cite{Zheng2017} used a Deep Convolutional GAN (DCGAN) together with the LSR algorithm to generate images and label them. They called this method Label Smooth Regularization for Outliers (LSRO) and it assigns the artificially generated images the same value in all classes, that is, it is uniformly distributed throughout all classes. The same year, Ainam et al. \cite{Ainam2019} proposed the use of the k-means algorithm to group the images for the training set of a DCGAN. They also introduced the Sparse Label Smoothing Regularization (SLSR) , which is based on the LSRO labeling method. It  labels the artificially generated images using the groups generated by k-means.

Interpolation of images has been also successfully tried. Eom et al. \cite{Eom2019} proposed a new GAN architecture called Identity Shuffle GAN (IS-GAN). In this method, the generation of artificial images is through of the interpolation between two real images, being able to distinguish between the upper and the lower part. The labels of the artificial image will be the same as the images that have generated the interpolation.

More recently, in 2021, Hussin et al. 2021 \cite{Hussin2021} proposed the use of the StyleGAN \cite{Karras2021} for the generation of new data. First, StyleGAN is trained with some of the most popular datasets for people re-identification. Once StyleGAN has been trained, a set of new artificial images is generated. Finally, the generated images are labelled using the LSRO method.

\section{DISCUSSION}
Data augmentation is a difficult task in any domain and the re-identification domain is not the exception. In this article we have reviewed the most relevant approaches for data augmentation in person re-identification systems which employ generative adversarial networks. We have categorized all the reviewed methods in three groups: style transfer, pose transfer, and random generation of images. 

Neither of these approaches can be regarded as the best one. The kind of application may indicate the use of one approach over the other. For example, a re-identification system operating inside a shop with controlled light conditions might require a data augmentation based on pose transfer, meanwhile an re-id application for outdoors surveillance might benefit more from style transfer models. In Table \ref{summaryGans} we provide the summary of all the methods here reviewed.

\begin{table}
\centering
\begin{tabular}{|c|c|c|c|c|}
\hline
\bf{GAN Model} & \bf{Approach} \\  
\hline	\hline 
CycleGAN \cite{Zhu2017}  & Style transfer \\
\hline
CamStyle \cite{Zhong2019}  & Style transfer  \\
\hline
cmGAN \cite{Dai2018a}  & Style transfer  \\
\hline
M2M-GAN \cite{Liang2018}  & Style transfer \\
\hline
DG-NET \cite{Zheng2019}  & Style transfer  \\
\hline
TC-GAN + DFE-Net \cite{Pang2021}  & Style transfer  \\
\hline
\hline
PN-GAN \cite{Qian2018} & Pose transfer  \\
\hline
WF+WPR \cite{Borgia2019} & Pose transfer  \\
\hline
PAC-GAN \cite{Zhang2020}  & Pose transfer  \\
\hline
GAN + CNN + STN \cite{Ni2021} & Pose transfer  \\
\hline \hline
LSR \cite{Szegedy2016} & Random generation  \\
\hline
DCGAN + LSR \cite{Zheng2017} & Random generation  \\
\hline
k-means + DCGAN SLSR \cite{Ainam2019} & Random generation  \\
\hline
IS-GAN \cite{Eom2019} & Random generation  \\
\hline
StyleGAN \cite{Karras2021} \cite{Hussin2021} & Random generation \\
\hline
\end{tabular} 
\caption{Summary of GAN approaches to data augmentation in person re-identification.} 
\label{summaryGans}
\end{table}

Moreover, even though good advances have been achieved so far with these approaches to data augmentation, we must keep in mind that generative adversarial networks are not trivial models to train and fine-tune. Common problems that researchers generally encounter when training GANs, such as mode collapse and training instability, are also very likely to appear when learning to augment a re-identification dataset.

Mode collapse occurs when the generator network becomes less and less diverse as the number training iterations increases. Basically, the generator improves the quality of one specific category of fake images that it produces, for example, women images, and begins to produce almost exclusively women images. Consequently, the discriminator network begins to experience only fake images of women and it also starts to forget to discriminate other category of images, for example, men images. With time, when the discriminator manages to discriminate images of women, the generator is forced to start producing another category of images, for example children images. This problem causes the GAN to cycle around a few number of categories, never becoming good in all the required categories.

The training instability problem has to do with the mere fact of training two neural networks at the same time, in a way that the good performance of one network depends on the good performance of the other.  In such training dependency, it is very easy for the whole system to get trapped into an unstable phase, in which all network's parameters oscillate. Very often, the training process might seem to be going very well and then, from one iteration to the other, the performance drops unexpectedly, without no clear reason. Under these circumstances, a lot of fine-tuning is necessary, in order to find an optimal set of hyperparameters.

\begin{figure*}
\centering
\includegraphics[width=0.5\linewidth]{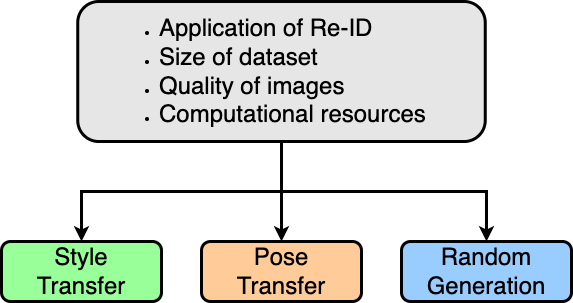} 
\caption{The selection of the best approach depends on several conditions such as the application of the Re-ID system, the size of the dataset, the quality of the images and the computational resources available.}
\label{diagram_approaches}
\end{figure*}

Regarding the quality of each approach, rather than saying that one method is better that the other, we provide a list of advantages and disadvantages in Tables \ref{summaryAdvantagesStyle}, \ref{summaryAdvantagesPose} and \ref{summaryAdvantagesRandom}, for style transfer, pose transfer and random generation, respectively. The selection of one approach over the other depends on several factors. Some of the main factors include: the specific application of the re-identification system, the size of the dataset, the resolution of the images, and the computational power available for training the models (see Fig. \ref{diagram_approaches}).

\begin{table}
\centering
\resizebox{\textwidth}{!}{%
\begin{tabular}{|c|c|c|}
\hline
\bf{GAN Model} &  \bf{Advantage} & \bf{Disadvantage} \\  
\hline	\hline 
CycleGAN \cite{Zhu2017}  & \makecell[l]{Image paired examples are not \\ needed.} & \makecell[l]{Results are not uniformly positive. \\ It works better for translations \\ involving color and texture than \\ geometric changes.} \\
\hline
CamStyle \cite{Zhong2019}  & \makecell[l]{It reduces the risk of deep network \\ overfitting and that smooths  \\ the CamStyle disparities.}  & \makecell[l]{It cannot directly handle multi-domain \\ image-to-image translation methods.}\\
\hline
cmGAN \cite{Dai2018a}  & \makecell[l]{It focuses on Re-ID between \\  infrared and RGB images and it  \\ can be trained in an end-to-end \\ manner.} & \makecell[l]{It was tested only for infrared-RGB \\ translations.}  \\
\hline
M2M-GAN \cite{Liang2018}  & \makecell[l]{It solves the many-to-many \\ cross-domain transfer learning \\ with less training time, fewer \\ parameters and a better performance.} & \makecell[l]{For every training example image, \\ it needs the source and target \\ sub-domain labels. This could \\ represent a problem if we do not \\ have those labels.} \\
\hline
DG-NET \cite{Zheng2019}  & \makecell[l]{It separately encodes each person \\ into  an appearance code and a \\ structure code.} & \makecell[l]{The generative module tends to learn \\ regular textures such as stripes and \\ dots, but it ignores some rare patterns \\ such as logos on shirts.} \\
\hline
\makecell{TC-GAN \\ + DFE-Net \cite{Pang2021}}  & \makecell[l]{It effectively combines supervised \\ and unsupervised learning.} & \makecell[l]{Training this model can take longer \\ than some of the methods studied in \\ their comparative work.} \\
\hline
\end{tabular} }
\caption{Summary of advantages and disadvantages of style transfer approaches.} 
\label{summaryAdvantagesStyle}
\end{table}

\begin{table}
\centering
\resizebox{\textwidth}{!}{%
\begin{tabular}{|c|c|c|}
\hline
\bf{GAN Model} &  \bf{Advantage} & \bf{Disadvantage} \\  
\hline	\hline 
PN-GAN \cite{Qian2018} & \makecell[l]{It can generalize to re-id datasets \\ collected  from  new  cameras without \\ model fine-tuning.} & \makecell[l]{In a transfer learning setting, \\ the reported results are lower than \\ those in a supervised setting, \\ which is expected due to the \\ complexity of the task.} \\
\hline
WF+WPR \cite{Borgia2019} & \makecell[l]{It mitigates the effects of the \\ inter-camera viewpoint problem.} & \makecell[l]{It relies on a network with a depth \\ of around 1/3 of ResNet50, which \\ could reduce its performance.} \\
\hline
PAC-GAN \cite{Zhang2020}  & \makecell[l]{It does not require labeled data \\ for training.} & \makecell[l]{The computation needed for generating \\ a larger number of authentic samples \\ is very expensive.} \\
\hline
\makecell{GAN + CNN \\ + STN \cite{Ni2021}} & \makecell[l]{It learns features containing global \\ information as well as local detailed \\ properties.} & \makecell[l]{The least performance improvement \\ occurs with the region of the legs \\
and feet of a person.}\\
\hline
\end{tabular} }
\caption{Summary of advantages and disadvantages of pose transfer approaches.} 
\label{summaryAdvantagesPose}
\end{table}

\begin{table}
\centering
\resizebox{\textwidth}{!}{%
\begin{tabular}{|c|c|c|}
\hline
\bf{GAN Model} &  \bf{Advantage} & \bf{Disadvantage} \\  
\hline	\hline 
LSR \cite{Szegedy2016} & \makecell[l]{It prevents the largest logit in \\ convolutional networks from becoming \\ much larger than all others.} & \makecell[l]{It is a general-purpose method for \\ training convolutional networks and \\ it was not specifically designed \\ for data augmentation.} \\
\hline
DCGAN + LSR \cite{Zheng2017} & \makecell[l]{It is complementary to previous \\ methods due to the regularization \\ of the GAN data.} & \makecell[l]{This method is more focused on the \\ regularization ability of the GAN than \\ on producing a state-of-the-art result.} \\
\hline
\makecell{k-means + \\ DCGAN SLSR \cite{Ainam2019}}  & \makecell[l]{It exploits the clustering property \\ of person
re-id datasets and creates \\ groups of similar objects in order \\ to model cross-view variations.} & \makecell[l]{It needs thousands of labels to better \\ handle a practical environment \\ scenario.} \\
\hline
IS-GAN \cite{Eom2019} & \makecell[l]{It disentangles identity-related and \\ indentity-unrelated features from \\ person images.} & \makecell[l]{This method does not have a particular \\ reported disadvantage, other than \\ ignoring features related to the \\ persons' identities.} \\
\hline
StyleGAN \cite{Karras2021} \cite{Hussin2021} & \makecell[l]{With high-resolution images, the \\ success rate increases.} & \makecell[l]{StyleGAN requires huge computational \\ resources for training.} \\
\hline
\end{tabular} }
\caption{Summary of advantages and disadvantages of random generation approaches.} 
\label{summaryAdvantagesRandom}
\end{table}

\section{CONCLUSION}

In this article we have reviewed three of the most common approaches for data augmentation in person re-identification systems which employ generative adversarial networks for style transfer, pose transfer, and random generation of images. We have included in this review, what we consider to be the most relevant and successful ways to augment image datasets for the task of re-identifying people. This is of course not a  exhaustive survey, but a reference guide to researchers or engineers interested in this field.

Finally, although data augmentation for re-identification systems is nowadays possible through the use of generative adversarial networks, using the approaches above mentioned, it is important to consider that this task may require an important amount of expertise in implementating, training and fine-tuning generative adversarial models.

\pagebreak 

\bibliographystyle{abbrv}
\bibliography{Referencias.bib}

\end{document}